\documentclass{article}
\usepackage{spconf,amsmath,graphicx,url,float,booktabs
,makecell,cite,color,array}

\title{AIGCOIQA2024: Perceptual Quality Assessment of AI Generated Omnidirectional Images}

\name{\parbox{\linewidth}{\centering \small Liu Yang$^1$, Huiyu Duan$^1$, Long Teng$^1$, Yucheng Zhu$^1$, Xiaohong Liu$^1$, Menghan Hu$^2$, Xiongkuo Min$^1$, Guangtao Zhai$^1$, Patrick Le Callet$^3$}}
\address{\parbox{\linewidth}{\centering \small $^1$ 
Shanghai Jiao Tong University, Shanghai, China\qquad 
$^2$ East China Normal University, Shanghai, China\\
$^3$ Université de Nantes, Nantes, France}\\
\small Email:\{ylyl.yl, huiyuduan, tenglong, zyc420, xiaohongliu, minxiongkuo, zhaiguangtao\}@sjtu.edu.cn,\\
\small mhhu@ce.ecnu.edu.cn, patrick.lecallet@univ-nantes.fr}

\begin{document}

\maketitle
\begin{abstract}
In recent years, the rapid advancement of Artificial Intelligence Generated Content (AIGC) has attracted widespread attention. Among the AIGC, AI generated omnidirectional images hold significant potential for Virtual Reality (VR) and Augmented Reality (AR) applications, hence omnidirectional AIGC techniques have also been widely studied. AI-generated omnidirectional images exhibit unique distortions compared to natural omnidirectional images, however, there is no dedicated Image Quality Assessment (IQA) criteria for assessing them. This study addresses this gap by establishing a large-scale AI generated omnidirectional image IQA database named AIGCOIQA2024 and constructing a comprehensive benchmark. We first generate 300 omnidirectional images based on 5 AIGC models utilizing 25 text prompts. A subjective IQA experiment is conducted subsequently to assess human visual preferences from three perspectives including quality, comfortability, and correspondence. Finally, we conduct a benchmark experiment to evaluate the performance of state-of-the-art IQA models on our database.
The database will be released to facilitate future research.
\end{abstract}
\vspace{-10pt}
\section{Introduction}
\label{sec:intro}
\vspace{-6pt}

AI Generated Content (AIGC) refers to generate various forms of content such as texts, images, musics, videos, and 3D interactive contents, \textit{etc}, using AI.
Thanks to the rapid advancement of generative models such as Generative Adversarial Network (GAN) \cite{gan}, Variational Auto Encoders (VAE) \cite{vae} and Diffusion Models (DMs) \cite{dm}, \textit{etc}., as well as langurage-vision models such as BLIP2 \cite{blip2} and CLIP \cite{clip}, \textit{etc.}, recent AI Generated Images (AIGIs) \cite{dalle, sd} have shown excellent performance, and have been extended to omnidirectional image generation. Omnidirectional images can provide 360-degree free viewing experience, which are widely used in Virtual Reality (VR), Augmented Reality (AR), game development, cultural heritage protection and other fields, which are generally shown in the Equirectangular Projection (ERP) format to warp the image and adapt to omnidirectional characteristics. The recent emergence of AI based omnidirectional image generation models \cite{mvdiffusion, text2light} have shown their potential of fast and creative omnidirectional content generation.

However, many omnidirectional images generated by AI are of low quality and cannot meet the visual expectations of the users. Compared with natural omnidirectional images, AI generated omnidirectional images may not only exhibit the issues such as blurriness, noise, distortion, and lighting problems , but also exist some unique degradations caused by the AI generation, such as unrealism, unreasonable composition, and low relevance between text and image. These quality degradation issues can significantly affect the immersive experience of the users in real-world applications. Therefore, it is significant and necessary to identify and evaluate the generated omnidirectional images that do not meet the user preferences, and either discard or modify them accordingly.

\begin{figure}[t]
\vspace{-3pt}
\begin{minipage}[b]{1.0\linewidth}
  \centering
  \centerline{\includegraphics[width=7.5cm]{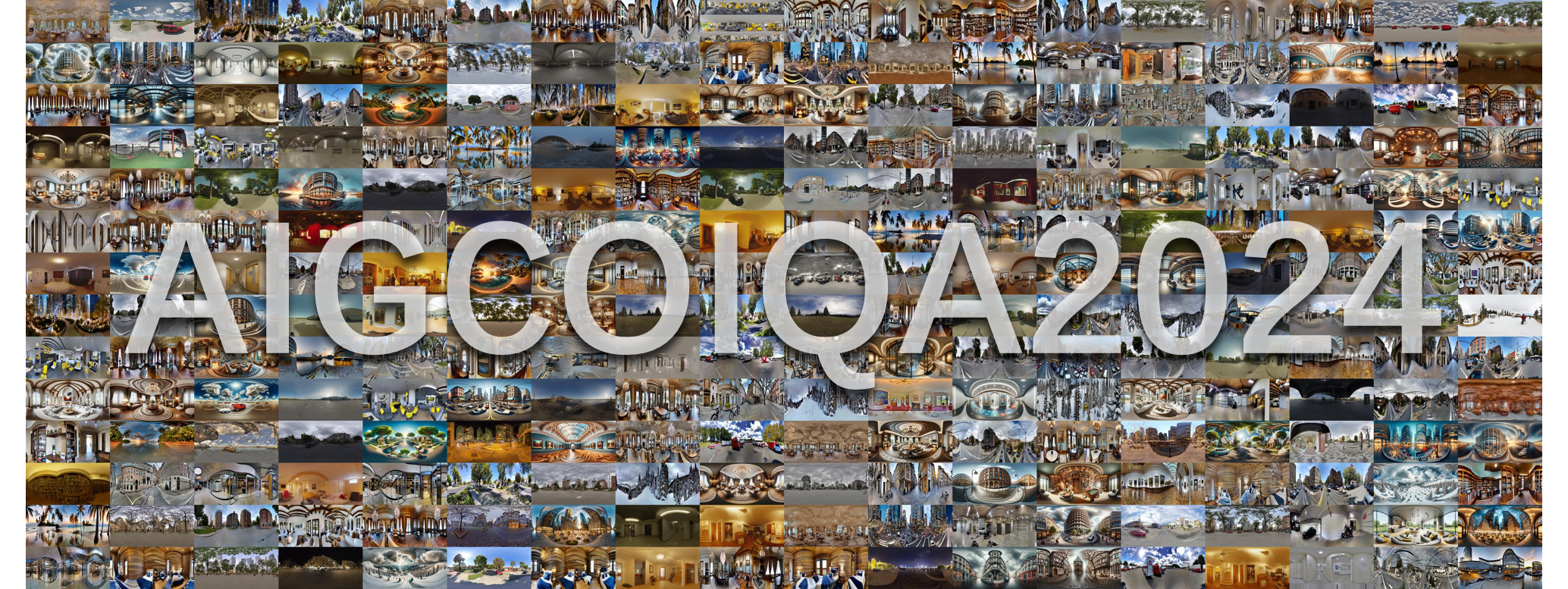}}
\end{minipage}
\vspace{-20pt}
\caption{An overview of the AIGCOIQA2024 database.}
\vspace{-13pt}
\label{fig:overview}
\end{figure}

Existing Image Quality Assessment (IQA) models primarily focus on low-level aspects of image quality, such as lighting, color, and clarity. As aforementioned, AI generated omnidirectional images may not only show degradations in low-level visual aspects but also in high-level preference aspects. While there are quantitative metrics such as Fréchet Inception Distance (FID) \cite{fid} proposed to evaluate the performance of generation models, these algorithms can only assess the authenticity dimension of an image set, which lack the capability to evaluate a single image and perform text-image correspondence measure. Other algorithms, like CLIPScore \cite{clipscore}, mainly consider the text-image correspondence dimension of AI-generated images, while ignoring the comprehensive assessment of authenticity and overall quality of omnidirectional images. Recently, some studies have established IQA databases for classical AI-generated images and performed benchmark experiments \cite{aigciqa}\cite{li2023agiqa}. 
However, IQA studies for AI-generated omnidirectional images are still lacking, which have significant difference compared with classical image in terms of  formats, characteristics, and applications.

To better understand human visual preferences for AI generated omnidirectional images and facilitate the development of IQA algorithms for such images, we construct a database termed AIGCOIQA2024, which contains 300 omnidirectional images and collected corresponding human preference ratings from three perspectives. Specifically, we first generate 300 omnidirectional images based on 5 different models with 25 text prompts describing diverse indoor and outdoor scenes. Different from general AIGC IQA databases, such as AIGCIQA2023  \cite{aigciqa}, in the subsequent subjective preference assessment experiment, subjects are instructed to score the images from the perspectives of quality, comfortability, and correspondence based on human visual preferences on account of the AIGC degradations and the applications of the omnidirectional images. Our contributions can be summarized as follows:
\vspace{-5pt}
\begin{itemize}
\item[$\bullet$]We propose to assess AI generated omnidirectional images from the perspectives of quality, comfortability, and correspondence to quantify human visual preferences.
\vspace{-5pt}
\item[$\bullet$]A human preference assessment database for omnidirectional images, termed AIGCOIQA2024, is established, which is the first AIGC IQA database for omnidirectional images to the best of our knowledge.
\vspace{-5pt}
\item[$\bullet$]We analyze the human preference characteristics for AI generated omnidirectional images based on the constructed database.
\vspace{-5pt}
\item[$\bullet$]We conduct a benchmark experiment, evaluating performance of some state-of-the-art (SOTA) models on our AIGCOIQA2024 database in terms of quality, comfortability, and correspondence.
\end{itemize}
\vspace{-3pt}

The remaining content of this paper is outlined as follows. In section 2, we describe the construction procedure of our AIGCOIQA2024 database in detail, including the process of image generation and the execution of subjective experiments. In section 3, we conduct statistics analysis for images in AIGCOIQA2024 along with the analysis of the subjective data. Section 4 introduces the benchmark experiment, showing the performance of state-of-the-art (SOTA) models on  AIGCOIQA2024 database. In Section 5 we conclude the paper and highlight future research directions.

\vspace{-4pt}
\section{AIGCOIQA2024 Database Construction}
\label{sec:AIGCOIQA2024 Database Construction}

\vspace{-4pt}
\subsection{Omnidirectional Image Generation}
\label{ssec:Omnidirectional Image Generation}
\vspace{-3pt}

In order to better understand human visual preferences for AI-generated omnidirectional images, we establish an IQA database containing 300 omnidirectional images. We first collect 25 omnidirectional images from the MVDiffusion \cite{mvdiffusion} \cite{matterport} training dataset and SUN360 \cite{sun360} as natural instances. Then we use BLIP2 \cite{blip2} to annotate the images to form preliminary prompts, and use GPT4 to add details and polish them. The generated prompts describe 12 indoor and 13 outdoor scenes, respectively, which are diverse and cover a wide range of scenarios. The descriptions of the prompts contain abundant scene details, making the generated images more distinguishable in terms of text-image correspondence.

\vspace{0pt}
For each prompt, we adopt 5 AIGC models, including MVDiffusion \cite{mvdiffusion}, Text2Light \cite{text2light}, DALLE \cite{dalle}, omni-inpainting \cite{mvdiffusion}\cite{inpainting} and a fine-tuned Stable Diffusion model \cite{sd} to generate omnidirectional images. When fine-tuning Stable Diffusion model \cite{sd}, we use 2000 and 4000 BLIP2-labeled \cite{blip2} indoor/outdoor ERP omnidirectional images to fine-tune the Unet module only to achieve indoor/outdoor omnidirectional generation. We generate two omnidirectional images for each of the first four generation models  mentioned above, respectively, and one for fine-tuned Stable Diffusion model \cite{sd}. For MVDiffusion \cite{mvdiffusion} generation, we adjust the denoising time to generate distinguishable images. Overall, taking natural omnidirectional images into account, we have a total of 12 omnidirectional images for each prompt, and a total of $(12+13)\times 12 = 300$ omnidirectional images in the database.
Fig. \ref{fig:overview} gives an overview of the images in our AIGCOIQA2024.

\begin{figure}[!t]
\vspace{-3pt}
\begin{minipage}[b]{1.0\linewidth}
  \centering
  \centerline{\includegraphics[width=7.5cm]{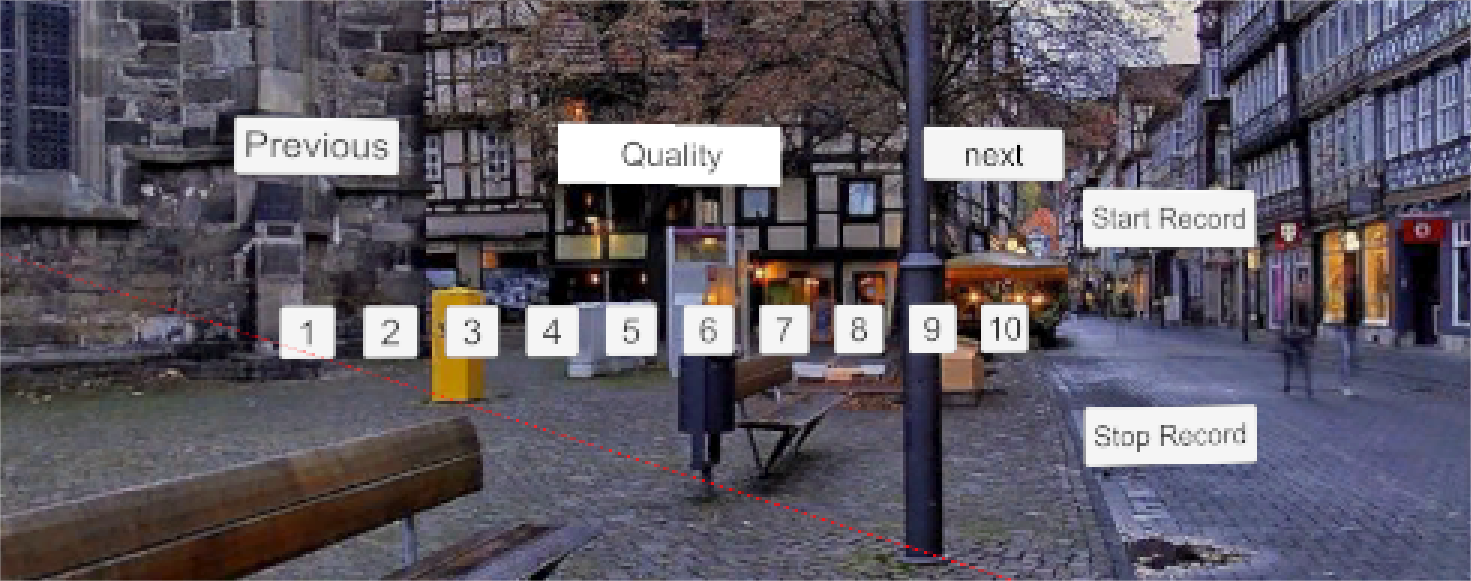}}
\vspace{-3pt}
\end{minipage}
\vspace{-18pt}
\caption{An example of the subjective experiment interface in unity, participants can use the cursor to click on the score box to select the ``quality'' score.
}
\vspace{-9pt}
\label{fig:interface}
\end{figure}

\begin{figure*}[!t]
\vspace{-4pt}
\begin{minipage}[b]{1.0\linewidth}
  \centering
  \centerline{\includegraphics[width=17.0cm]{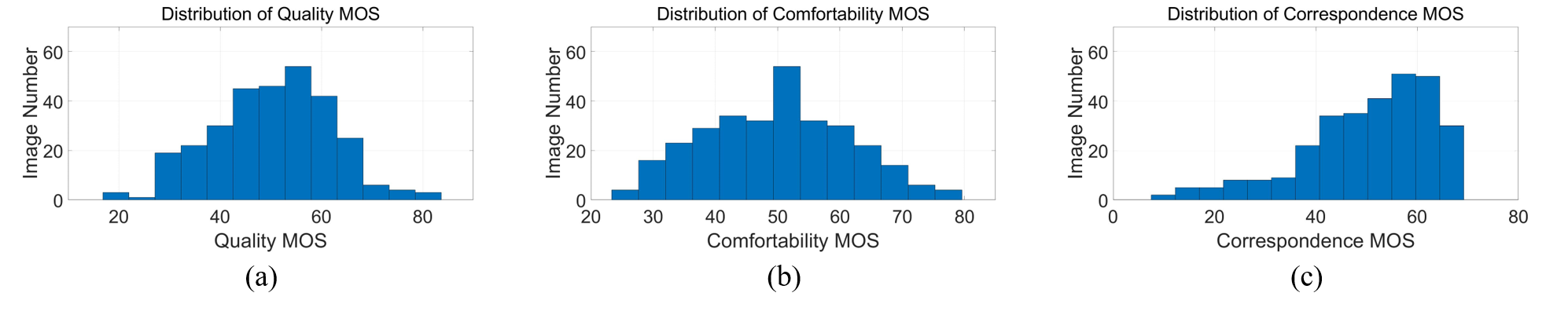}}
\end{minipage}
\vspace{-30pt}
\caption{(a) MOS distribution of quality score.(b) MOS distribution of comfortability score.(c) MOS distribution of correspondence score.}
\vspace{-3pt}
\label{fig:data}
\end{figure*}
\begin{figure*}[!t]
\begin{minipage}[b]{1.0\linewidth}
  \centering
  \centerline{\includegraphics[width=17.0cm]{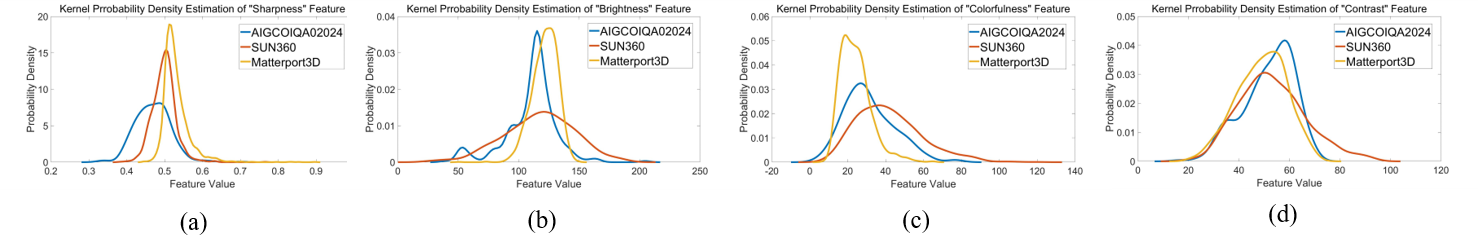}}
\end{minipage}
\vspace{-30pt}
\caption{Kernel distribution of four selected features of three databases: AIGCOIQA2024, SUN360 \cite{sun360}, Matterport3D \cite{matterport}.}
\vspace{-10pt}
\label{fig:kernel}
\end{figure*}
\begin{figure}[!t]

\begin{minipage}[b]{1.0\linewidth}
  \centering
  \centerline{\includegraphics[width=8.5cm]{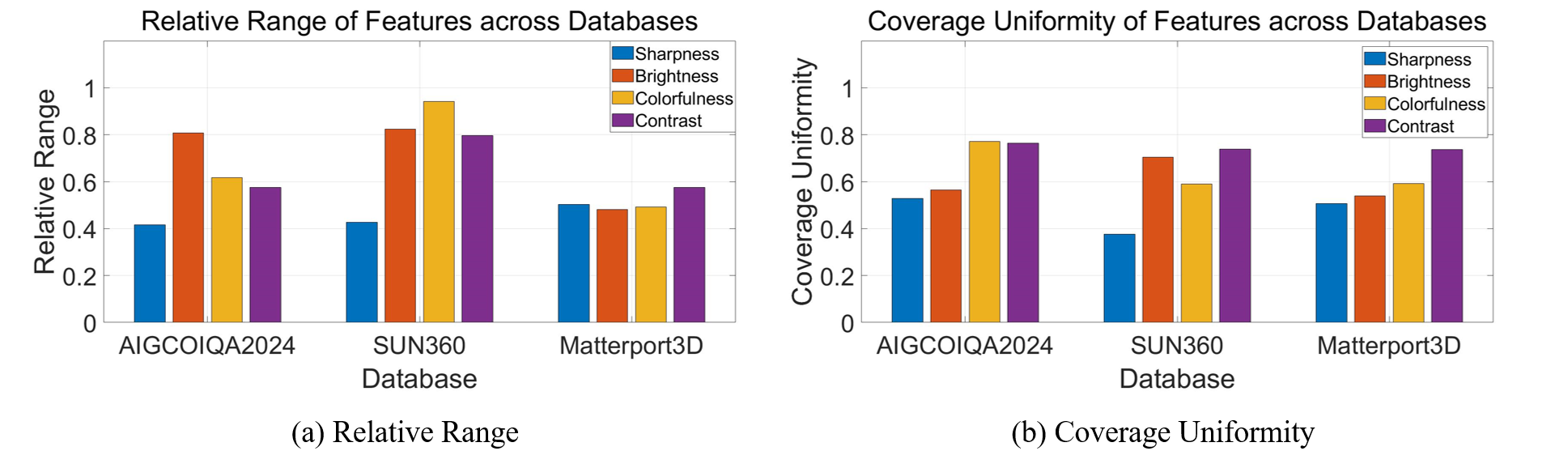}}
\end{minipage}
\vspace{-25pt}
\caption{Relative range and Coverage uniformity of the four selected features on the three databases.}
\vspace{-11pt}
\label{fig:U}
\end{figure}

\vspace{-3pt}
\subsection{Subjective Experiment Setup}
\label{ssec:subjective experiment setup}
\vspace{-3pt}
To measure human visual preferences for generated omnidirectional images, we further conduct a subjective evaluation experiment. Due to the inherent characteristics of omnidirectional images generated by AI, we cannot measure human visual preferences only from the dimension of "quality", as discussed in AIGCIQA2023 \cite{aigciqa}. However, different from the general AIGC IQA problem \cite{aigciqa}, AI generated omnidirectional images are mainly produced for VR and AR applications, in which users generally have particular visual characteristics \cite{ref1,ref2,ref3,ref4,ref5,ref6,ref7,ref8}. Therefore, in this paper, we propose to evaluate human visual preferences for generated omnidirectional images from three perspectives, including quality, comfortability and correspondence.

The first evaluative dimension is ``\textbf{quality}'', \textit{i.e.} a comprehensive score of low-level visual qualities such as color, lighting and clarity, \textit{etc}. Since generated omnidirectional images are mainly viewed in VR environments, We further present the second evaluation perspective, \textit{i.e.,} ``\textbf{comfortability}'', which is defined as the user experience preference for AI-generated omnidirectional images. Specifically, subjects are asked to give an overall score for the authenticity level, structure deformation level, as well as the comfort level.
Since the generated omnidirectional images are mainly produced via the control of the text prompts, the correspondence between text and image is also a critical criteria for evaluating the quality of AI-generated omnidirectional images, \textit{i.e.} ``text-image \textbf{correspondence}''.

\vspace{0pt}
Then we conduct a subjective experiment under the guidance of ITU-R BT.500-14 \cite{itu}, with a total of 20 subjects participating (10 males and 10 females). All subjects have normal or corrected-to-normal vision. After browsing 13 training samples, the participants are asked to view 300 omnidirectional images and score them from the perspectives of quality, comfortability and correspondence ranging from 1 to 10 with an interval of 1 based on the subjecinterfative perception. The images are randomly sorted and displayed sequentially in head-mounted displays (HMDs) based on the software designed using the Unity, as shown in Fig. \ref{fig:interface}.

\vspace{-9pt}
\subsection{Subjective Data Processing}
\label{ssec:subjective data Processing}
\vspace{0pt}
We follow the instructions of ITU \cite{ITU-R_BT_500-13} to conduct the outlier detection and subject rejection. As a result, no subjects are rejected and the rejection ratio is 3$\%$ for all ratings. For the remaining valid subjective scores, we convert the raw ratings into Z-scores, then linearly scale them to the [0,100] range. The final 
Mean Opinion Score (MOS) is calculated as follows: 
\vspace{-6pt}
\begin{gather}
z_{ij}=\frac{m_{ij}-\mu_{i}}{\sigma_{i}},\quad z'_{ij}=\frac{100z_{ij}+3}{6}\\
MOS_{j}=\frac{1}{N}\sum_{i=1}^{N}z'_{ij}
\end{gather}
where $m_{ij}$ is the subjective score given by the $i$-th subject to the $j$-th image, $\mu$ is the mean score given by the $i$-th subject, $\sigma$ is the standard deviation and $N$ is the total number of subjects.

Fig. \ref{fig:data} illustrates histograms of the MOSs of quality, comfortability, and correspondence perspectively. The MOSs distribution shows that our database encompasses a broad range of scores, indicating its diversity. Moreover, different perspectives have different distributions, which also illustrates the differences between the three evaluation perspectives.
\vspace{-6pt}
\section{Database Analysis}
\label{sec:Database analysis}
\vspace{0pt}
\subsection{Statistics Analysis for Generated Omnidirectional Images}

\begin{figure*}[!t]
\centering
\includegraphics[width=0.95\textwidth]{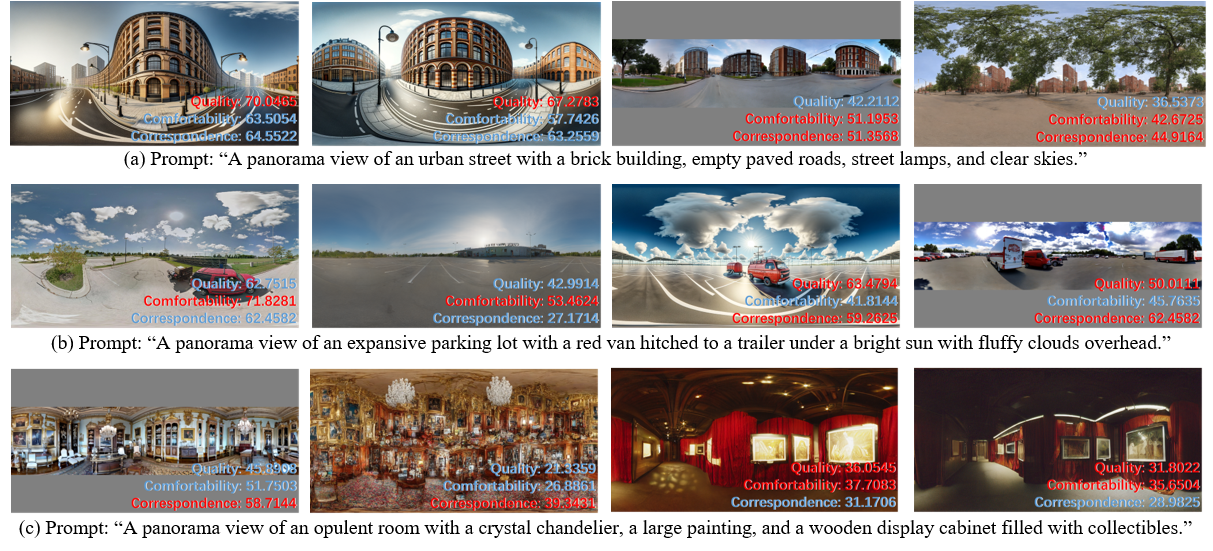}
\vspace{-5pt}
\caption{Comparison of the differences between three evaluation perspectives. (a) The left two omnidirectional images have better quality, but worse comfortability and correspondence. (b) The left two omnidirectional images have better comfortability, but worse quality and correspondence. (c) The left two omnidirectional images have better correspondence, but worse quality and comfortability.}
\vspace{-9pt}
\label{fig:comparison}
\end{figure*}
We conduct statistical analysis for our AIGCOIQA2024 database in terms of four low-level vision feature dimensions including: ``sharpness'', ``brightness'', ``colorfulness'' and ``contrast''. Some images from real-world omnidirectional databases, \textit{i.e.}.  Matterport3D \cite{matterport} and SUN360 \cite{sun360} for comparison. For simplicity, we refer to these four features as ${Ci}, (i = 1, 2, 3, 4)$, respectively. Fig. \ref{fig:kernel} shows the kernel distribution of each feature for these three databases. It can be observed that the generated omnidirectional images have a wide distribution in ``sharpness'' and ``colorfulness'' features, showing their diversity. For the ``contrast'' feature, generated omnidirectional images show similar characteristics compared with other two natural omnidirectional databases.
However, for the ``brightness'' feature, our database shows narrow distribution range compared to SUN360 \cite{sun360}.

\begin{table*}[!t]
\centering
\vspace{-3pt}
\caption{Performance comparision of the state-of-the-art NR-IQA models on the evaluation of human preferences for AI generated omnidirectional images from the perspectives of quality, comfortability and correspondence. The best performances are marked in \textcolor{red}{RED} and the second-best performances are marked in \textcolor{blue}{BLUE}.}
\vspace{2pt}
\label{1}
\setlength{\tabcolsep}{1.3em}
\scalebox{0.83}{
\begin{tabular}{l|c c c|c c c|c c c}
\toprule
Dimension & \multicolumn{3}{c|}{Quality} & \multicolumn{3}{c|}{Comfortability} & \multicolumn{3}{c}{Correspondence}\\
\midrule
Model&SRCC&KRCC&PLCC&SRCC&KRCC&PLCC&SRCC&KRCC&PLCC\\
\midrule
QAC \cite{QAC}&0.1802&0.1197&0.0933&0.2665&0.1823&0.2483&0.1756&0.1175&0.0023\\
BMPRI \cite{bmpri}&0.4285&0.3010&0.5612&0.2652&0.1776&0.3728&0.2705
&0.1780&0.3092\\
NIQE \cite{niqe}&0.6858&0.4814&0.6236&0.5741&0.4128&0.6072&0.6672&0.4786&0.5776\\
ILNIQE \cite{ilniqe}&0.0434&0.0293&0.0617&0.0298&0.0186&0.0902&0.2959&0.1964&	0.3055\\
HOSA \cite{hosa}&0.7114&0.5154&0.7175&0.4793&0.3282&0.5020&\textcolor{red}{0.7200}&\textcolor{blue}{0.5262}&	0.6738\\
BPRI-PSS \cite{pri}&0.3673&0.2465&0.4299	&0.2153&0.1427&0.2216&0.5782	&0.4101&0.6491\\
BPRI-LSSs \cite{pri}&0.3018&0.2094&0.4398&0.2435&0.1668&0.3516&0.1004&0.0672&
0.1536\\
BPRI-LSSn \cite{pri}&0.3604&0.2300&0.5095&0.1506&0.0825&0.3268&0.5532&0.3878&
0.5433\\
BPRI \cite{pri}&0.3553&0.2503&0.4849&0.2866&0.1986&0.4092&0.1863&0.1269&0.2319\\
FISBLIM \cite{fisblim}&0.6472&0.4588&0.6124&0.4425&	0.3015&0.3492&0.6836&0.4929&0.5493\\
\midrule
CLIPScore \cite{clipscore}&0.3308&0.2241&0.3320&0.1752&0.1192&0.1666&0.3809&0.2641&0.4915\\
\midrule
CNNIQA \cite{cnniqa}&0.7066&0.5127&0.6345&0.5715&0.4068&0.5709&0.5935&0.4222&0.5291\\
Resnet18 \cite{resnet}&0.7722&0.5852&0.7286&0.6537&0.4728&0.6047&0.6709&0.4801&0.6829 \\
Resnet34 \cite{resnet}&0.6343&0.4632&0.6478&0.6551&0.4747&0.6177&0.5414&0.3848&0.6407 \\
VGG16 \cite{vgg}&0.7956&0.6047&0.7351&0.7074&0.5309&0.6616&0.6598&0.4747&0.6936\\
VGG19 \cite{vgg}&0.7628&0.5731&0.7105&0.7310&0.5439&0.6893&0.6773&0.4927&\textcolor{blue}{0.7054}\\
HyperIQA \cite{hyperiqa}&\textcolor{blue}{0.8354}&\textcolor{red}{0.6405}&\textcolor{blue}{0.7769}&0.7477&0.5564&0.7516&0.6506&0.4701&0.7052\\
MANIQA \cite{maniqa}&0.8038&0.6160&0.7682&\textcolor{red}{0.7837}&\textcolor{red}{0.5906}&\textcolor{red}{0.7796}&\textcolor{blue}{0.7198}&\textcolor{red}{0.5303}&\textcolor{red}{0.7598} \\
TReS \cite{tres}&\textcolor{red}{0.8355}&\textcolor{blue}{0.6378}&\textcolor{red}{0.7803}&\textcolor{blue}{0.7768}&\textcolor{blue}{0.5857}&\textcolor{blue}{0.7763}&0.7036&0.5174&0.6996\\
\bottomrule
\end{tabular}
}
\vspace{-6pt}
\end{table*}

To evaluate the distribution and uniformity of the databases over the four features, we further compute and compare the relative range and uniformity of coverage.
The relative range is calculate as:
\vspace{-9pt}
\begin{gather}
R_i^{d} = \frac{max(C_i^{d})-min(C_i^{d})}{max_d(C_i^d)}
\end{gather}
where $C_i^{d}$ refers to the data distribution of database $d$ on feature $i$, and $max_d(C_i^d)$ refers to the maximum value on feature $i$ across all databases.
Uniformity of coverage is calculated as the entropy of the B-bin histogram of $C_i^d$ over all sources for each database $d$:
\vspace{-9pt}
\begin{gather}
U_i^{d} = -\sum_{b=1}^{B} p_{b}log_{B}p_{b}
\end{gather}
where $p_b$ is the normalized number of souces in bin $b$ for each feature in each database.

\vspace{0pt}
The relative range and uniformity of coverage are plotted in Fig. \ref{fig:U}, quantifying intra- and inter-database differences. It can be concluded from the figures that the images in our AIGCOIQA2024 dataset cover diverse ranges regarding the four selected features.
\vspace{-6pt}

\subsection{Preferences Analysis from Three Perspectives for Generated Omnidirectional Images}
\vspace{0pt}
In order to further emphasize the different focus of our three evaluation dimensions and to verify the necessity of evaluating these three dimensions separately for a single AI generated omnidirectional image, three sets of examples are given in Fig. \ref{fig:comparison}. The four images in each set are generated based on the same prompt.
For the left two images in each set, one dimension score is significantly higher than other two scores, while for the right two images, that score is significantly lower than other two scores. In Fig. \ref{fig:comparison} (a), Fig. \ref{fig:comparison} (b), Fig. \ref{fig:comparison} (c), the left two images have higher quality scores, comfortability scores, correspondence scores, respectively, while the right two images have lower scores in these corresponding dimensions.
We conclude that different rating perspectives can reflect different human preferences, which are related but distinct.
Therefore, the assessment for a generated omnidirectional image must be performed from multiple dimensions.

Summarizing from the above, IQA of AI-generated omnidirectional images from any of these three dimensions alone is one-sided, and these three evaluation dimensions are independent. Therefore, when evaluating IQA of AI-generated omnidirectional images, we must evaluate an image comprehensively from all the three dimensions of quality, comfortability, and correspondence proposed above.

\vspace{-6pt}
\section{Experiment}
\label{sec:Experiment}

\subsection{Benchmark Models}
\label{ssec:Benchmark Models}

To evaluate the performance of various existing models on the prediction of human visual preferences for AI generated omnidirectional images, we employ 19 state-of-the-art no-reference (NR) IQA models for comparison. The selected models can be categorized into three groups:
\vspace{-3pt}
\begin{itemize}
\item[$\bullet$]\textbf{Handcrafted-based} models, which include QAC \cite{QAC}, BMPRI \cite{bmpri}, NIQE \cite{niqe}, ILNIQE \cite{ilniqe}, HOSA \cite{hosa}, FISBLIM \cite{fisblim}, BPRI-PSS \cite{pri}, BPRI-LSSs \cite{pri}, BPRI-LSSn \cite{pri} and BPRI \cite{pri}.
\vspace{-6pt}
\item[$\bullet$]\textbf{Vision-language pretrained} models, which include CLIPScore \cite{clipscore}.
\vspace{-6pt}
\item[$\bullet$]\textbf{Deep learning-based} models, including CNNIQA \cite{cnniqa}, Resnet18 \cite{resnet}, Resnet34 \cite{resnet}, VGG16 \cite{vgg}, VGG19 \cite{vgg},  HyperIQA \cite{hyperiqa}, MANIQA \cite{maniqa} and TReS \cite{tres}.
\end{itemize}
\vspace{-3pt}

For traditional hand-crafted models, we directly employ them to predict the preference scores for the omnidirectional images in our database. For CLIPScores \cite{clipscore}, we simply calculate the score using the cosine similarity between text and image embeddings. For deep learning-based IQA models, we partition the database into training and testing sets with a ratio of 3:2, without scene or text-prompt repeat. based on different scenes. The training parameters are setthe same as the officially released code.
\vspace{-3pt}
\subsection{Performance Analysis}
\label{ssec:Performance Analysis}

Three evaluation metrics, including Spearman Rank Correlation Coefficient (SRCC), Pearson Linear Correlation Coefficient (PLCC), and Kendall’s Rank Correlation Coefficient (KRCC), are adopted to evaluate the performance of the  models from the perspectives of quality, comfortability, and correspondence.

Table \ref{1} demonstrates the performance of the aforementioned state-of-the-art models. It can be observed that, in general, hand-crafted models show poor performance for evaluating the human preference of generated omnidirectional images, and these models perform almost worse for the dimension of comfortability compared to other two dimensions.
Deep learning-based models generally outperform hand-crafted models, but their performance is still not entirely satisfactory. Particularly, they struggle to achieve good performance in quality, comfortability and correspondence simultaneously.
These models generally perform better in the quality dimension but worse in the comfortablity and correspondence dimensions, due to the unawareness of the authenticity and comfortable texture of natural omnidirectional images, as well as the ignoring of utilizing the text information.
These can be explored in future works.

\section{Conclusion and Future Works}
\label{sec:Conclusion and Future Works}
It is important to study the human preferences for AI-generated omnidirectional images, which is rarely researched in the current literature. To this end, we first construct a database named AIGCOIQA2024, AI-generated omnidirection images and corresponding collected preference ratings from the perspectives of quality, comfortability and correspondence, respectively. Based on the database, we analyze the human visual preference characteristics for the generated omnidirectional images, and conduct a benchmark study.
The current models cannot well handle this new task.
It is worthwhile to explore the characteristic of natural omnidirectional images to improve the comfortability evaluation and exploit text information to improve the correspondence assessment in the future.

\bibliographystyle{IEEEbib}
\bibliography{strings,refs}

\end{document}